\documentclass{article}
\usepackage{bbm, enumerate}
\usepackage[inline, shortlabels]{enumitem}
    \PassOptionsToPackage{round, authoryear}{natbib}
\usepackage{xr}

\makeatletter
\newcommand*{\addFileDependency}[1]{
  \typeout{(#1)}
  \@addtofilelist{#1}
  \IfFileExists{#1}{}{\typeout{No file #1.}}
}
\makeatother

\newcommand*{\myexternaldocument}[1]{%
    \externaldocument{#1}%
    \addFileDependency{#1.tex}%
    \addFileDependency{#1.aux}%
}

\myexternaldocument{neurips_2023_supp}

\usepackage[preprint]{neurips_2023}

\usepackage{algorithm}
\usepackage{algpseudocode}

\usepackage[utf8]{inputenc} 
\usepackage[T1]{fontenc}    
\usepackage{url}            
\usepackage{booktabs}       
\usepackage{amsfonts}       
\usepackage{nicefrac}       
\usepackage{microtype}      
\usepackage{xcolor}         
\usepackage{amsmath, amssymb, mathabx}
\usepackage{graphicx}
\usepackage{hyperref}       
\hypersetup{
    colorlinks=true,
    citecolor = blue
}
\usepackage{wrapfig}
\usepackage{lipsum}

\renewcommand{\hat}{\widehat}
\renewcommand{\tilde}{\widetilde}
\renewcommand{\check}{\widecheck}

\newcommand{\pr}{\mathbb{P}}

\usepackage{amsthm}
\theoremstyle{plain}
\newtheorem{theorem}{Theorem}[section]
\newtheorem{lemma}{Lemma}[section]

\newtheorem{definition}{Definition}[section]

\newtheorem{remark}{Remark}[section]
\newtheorem{assumption}{Assumption}

\definecolor{yuancolor1}{rgb}{0.56, 0.0, 1.0}
\definecolor{yuancolor2}{rgb}{0.0, 0.42, 0.24}
\definecolor{yuancolor3}{rgb}{0.0, 0.19, 0.33} 
\definecolor{yuancolor3}{rgb}{0.2, 0.2, 0.6} 
\definecolor{meijiacolor}{RGB}{119, 50, 168} 

\title{Distribution-Free Matrix Prediction Under Arbitrary Missing Pattern}

\author{%
  Meijia Shao
  \\
  Department of Statistics\\
  The Ohio State University\\
  Columbus, OH 43210\\
  \texttt{shao.390@osu.edu} \\
  \And
  Yuan Zhang\\
  Department of Statistics\\
  The Ohio State University\\
  Columbus, OH 43210\\
  \texttt{yzhanghf@stat.osu.edu}
}

\begin{document}

\maketitle

\begin{abstract}
    This paper studies the open problem of conformalized entry prediction in a row/column-exchangeable matrix.
    The matrix setting presents novel and unique challenges,
    but there exists little work on this interesting topic.
    We meticulously define the problem, differentiate it from closely related problems, and rigorously delineate the boundary between achievable and impossible goals.
    We then propose two practical algorithms. 
    The first method provides a fast emulation of the full conformal prediction, while the second method leverages the technique of algorithmic stability for acceleration.
    Both methods are
    computationally efficient and can effectively safeguard coverage validity in presence of arbitrary missing pattern.
    Further, we quantify the impact of missingness on prediction accuracy and establish fundamental limit results.
    Empirical evidence from synthetic and real-world data sets corroborates the superior performance of our proposed methods.
\end{abstract}


\section{Introduction}
\label{section::introduction}

\subsection{Problem set up}
\label{subsec::introduction::problem-set-up}

Consider a matrix $A \in \mathbb{R}^{m_1 \times m_2}$. We observe its entries with missing values, denoted by $M \in \{0, 1\}^{m_1 \times m_2}$. Here, $M_{i,j} = 0$ if we observe $A_{i,j}$, and $M_{i,j} = 1$ if $A_{i,j}$ is missing. 
The observed data consist of $M$ and $\{A_{i,j}: M_{i,j}=0\}$.
Throughout this paper, we simplify the presentation by assuming the following assumption -- we will discuss relaxations of this assumption in Section \ref{section::discussion}.
\begin{assumption}[Simplification]
    \label{Assumption::A-1::regularity-conditions}
    The matrix $A$ is symmetric and universally bounded, i.e., $m_1=m_2=n+1$; 
    $A_{i,j}=A_{j,i}$, (thus $M_{i,j}=M_{j,i}$), and $|A_{i,j}|\leq C_0, \forall i,j$ for a known constant $C_0>0$.
\end{assumption}

The matrix $A$, as described in Assumption \ref{Assumption::A-1::regularity-conditions}, can represent the adjacency matrix of a weighted, undirected network. 
We examine the scenario when a new node, indexed $n+1$, joins the network represented by $A_{[1:n],[1:n]}$. Our goal is to predict the connection between node $n+1$ and an existing node, such as node $n$, based on the remaining observed values in $A$, formally:

\begin{definition}[{\bf The Distribution-Free Matrix Prediction Problem}]
    The goal is to construct a prediction set ${\cal I}_{n,n+1}$ for $A_{n+1,n}$ that satisfies proper coverage validity, as will be discussed in Section \ref{subsec::understanding::three-types-validity}, under only weak regularity assumptions about the distribution of $A$ and without specifying the missing pattern in $M$ (except $M_{n+1,n}\equiv0$).
\end{definition}

\subsection{Conformal prediction in the classical setting}
\label{subsec::introduction::classical-conformal-prediction}

Conformal prediction \citep{vovk2005algorithmic} is a key tool in achieving \emph{distribution-free} prediction.
Now we briefly review its gist in the context of regression.
Consider i.i.d. $(X_1,Y_1),\ldots,(X_{n+1},Y_{n+1})\sim {\cal F}_{{\cal X}\times {\cal Y}}$,
where $Y_{n+1}$ is missing and we want to predict it for $X_{n+1}$.
The core idea is to use a prediction algorithm that treats input data points symmetrically, such that applying it to the complete data (including $(X_{n+1},Y_{n+1})$) yields an \emph{exchangeable} residual sequence ${\cal R}(Y_{n+1}) := \big(|Y_1 - \hat Y_1(Y_{n+1})|,\ldots,|Y_{n+1} - \hat Y_{n+1}(Y_{n+1})|\big)$, in which, switching any two elements will not alter the joint distribution.
This implies: $\pr(|Y_{n+1}-\hat Y_{n+1}(Y_{n+1})| \leq {\cal Q}_{1-\alpha}({\cal R}(Y_{n+1})))\geq 1-\alpha$, where ${\cal Q}_\beta(\cdot)$ denotes the lower-$\beta$ quantile operator \citep{vovk2005algorithmic}.
As a result, the \emph{conformal prediction set} ${\cal I}_{n+1}:=\big\{z: |z-\hat Y_{n+1}(z)| \leq {\cal Q}_{1-\alpha}({\cal R}(z))) \big\}$ guarantees $\pr(Y_{n+1}\in {\cal I}_{n+1})\geq 1-\alpha$.
The allure of conformal prediction lies in that its coverage guarantee always holds, even when the model fit to the data is wrong.
This magic occurs because an inaccurate prediction about the future will deviate in the same pattern as it did in the past.
By capturing and mimicking the estimator's past behavior, we can properly calibrate its future predictions to ensure coverage validity.
Readers can observe two immediate generalizations:
first, the data pairs $(X_i,Y_i)$ only need to be exchangeable, instead of i.i.d.;
second, the residual $|Y_i - \hat Y_i(z)|$ can be generalized into a \emph{non-conformity score} that measures the unusualness of $Y_i$. 
This score can be well-defined without using a fitted value $\hat Y_i$ as a proxy \citep{lei2014distribution}.

\subsection{The matrix setting}
\label{subsec::introduction::the-matrix-setting}

Introducing conformal prediction technique into the matrix setting is quite nontrivial.
At least two key distinctions separate this from the classical setting.
The first is \emph{different forms of exchangeability.}
In the classical setting, exchangeability refers to distribution invariance with \emph{any two} indexes swapped.
However, in the matrix setting, ``exchangeability'' typically means permuting rows and/or columns will not change the marginal distribution of $A$.
We formally state this as follows.
\begin{assumption}[Matrix exchangeability]
    \label{Assumption::A-2::matrix-exchangeability}
    Throughout this paper, $A$ is an order-$(n+1)$ submatrix from an infinite exchangeable array $A^{(\infty)}\in \mathbb{R}^{\mathbb{N}\times \mathbb{N}}$, that is, for any permutation $\pi: \mathbb{N}\leftrightarrow \mathbb{N}$, we have 
    $
        A^{(\infty)} 
        \stackrel{d}=
        A^{(\infty)}_{\pi, \pi},
    $
    where $(A^{(\infty)}_{\pi,\pi})_{i,j} := A^{(\infty)}_{\pi^{-1}(i),\pi^{-1}(j)}$ and ``$\stackrel{d}=$'' means equal in distribution.
\end{assumption}
Assumption \ref{Assumption::A-2::matrix-exchangeability} implies that $A$ admits the Aldous-Hoover representation \citep{kallenberg1989representation,zhao2015hypergraph} as follows:
$
    A_{i,j}|\xi_i,\xi_j
    \stackrel{\textrm{indep.}}\sim
    f(\xi_i,\xi_j; \cdot),
$
where latent $\xi_i\stackrel{\rm i.i.d.}\sim$~Uniform$(0,1)$, and $f(\xi_i,\xi_j; \cdot)$ is some distribution function.
Notably different from the matrix estimation literature \citet{chatterjee2015matrix, gao2015rate}, in this paper, we make {\bf no} smoothness or other structural assumption on $f$ \citep{choi2017co}.

The second difference is \emph{the absence of a natural predictor},  and the fact that \emph{any artificially constructed predictor is influenced by missing values}.
If we draw an analogy between matrix prediction and classical regression, we can naturally perceive $A_{i,j}$'s as \emph{responses}.
However, what statistics of the data can be sensibly considered as \emph{predictors}?
Furthermore, missing entries (that follow an unknown missing pattern) will almost certainly impact any predictors we construct artificially.
We shall carefully address this significant challenge in Section \ref{section::our-method}.

\subsection{Related work}

Matrix prediction is intimately related to \emph{matrix imputation}, also known as \emph{matrix completion} \citep{rudelson2007sampling, cai2010singular, candes2010matrix, keshavan2010matrix, koltchinskii2011nuclear, candes2011robust, chatterjee2015matrix, gao2016optimal, klopp2017robust, cherapanamjeri2017nearly}.
In Section \ref{subsec::understanding::inspection-error}, we will carefully clarify their connections and distinctions and justify why we have chosen matrix prediction as our goal.

Conformal prediction, first introduced by \citep{gammerman1998learning, vovk2005algorithmic}, has attracted significant research interest, particularly since early 2010s.
An incomplete list of methodological developments includes
\citep{papadopoulos2002inductive, lei2014distribution,lei2018distribution, romano2019conformalized,chernozhukov2021distributional, barber2021predictive, barber2022conformal}.
Applications across various fields have been explored in
\citet{lei2021conformal, vovk2021testing, chernozhukov2021exact, candes2023conformalized, hu2023distribution}.
Comprehensive tutorials and review articles, such \citep{shafer2008tutorial, angelopoulos2022conformal, angelopoulos2023conformal, fontana2023conformal}, provide up-to-date references.

On the other hand, to our best knowledge, there exists little study on conformal prediction for matrix data.
The most relevant works we could find are \citet{himabindu2018conformal, kagita2022inductive}, where the authors proposed heuristic-based procedures. 
However, without strong assumptions such as \emph{missing completely at random (MCAR)}, the validity of their methods yet remains to be clarified (cf. our Section \ref{subsec::understanding::inspection-error}).
A very recent work \citet{zaffran2023conformal} addressed conformal prediction with missing values in the regression setting.
A concurrent work \citet{gui2023conformalized} investigated a related yet distinct scenario where entries are missing independently of a \emph{given} $A$, potentially at non-uniform rates; in contrast, our study assumes $A$ is exchangeable but allows arbitrary missing patterns that may depend on $A$. 
Given the distinct settings of the two papers, their coverage validity guarantees are not directly comparable.

\subsection{Our contributions}
Our contributions are three-fold.
First, we defined the distribution-free matrix prediction problem and carefully differentiated it from the closely-related yet distinct problem of matrix imputation.
Through our impossibility theorems, we clarified what goals are achievable and what are not.
Intriguingly, in the matrix setting, we can formulate three types of coverage validity (see Section \ref{subsec::understanding::three-types-validity}), in contrast to only two types in the regression setting.

Second, we proposed a full conformal method for matrix prediction and proved its coverage validity.
The full conformal method, while comprehensive, is computationally intractable when multiple entries are missing. 
To address this, we proposed two algorithms that expedite computation in different ways.
The first method emulates full conformal by combining multiple initial guesses with a marginal full conformal.
The second method leverages algorithmic stability \citep{ndiaye2022stable} to reduce the dimensionality of full conformal.
Notably, our second algorithm illustrates that \citet{ndiaye2022stable}'s method is not merely a means of acceleration, but also a potent tool for managing missing data in conformal prediction, which is of independent research interest.
Both of our methods compute fast and can effectively handle arbitrary missingness.

Third, we quantified the impact of missingness on prediction accuracy both empirically, through numerical studies, and theoretically, through our fundamental limit theorem.

Our paper can be seen as one of the first efforts towards the development of theoretically guaranteed conformalized methods for matrix and network data. 
We anticipate many interesting future works to follow and believe that our paper will help open up a promising new line of research in this subfield.

\section{Understanding the problem}
\label{section::understanding-the-problem}


\subsection{Matrix prediction versus matrix imputation}
\label{subsec::understanding::inspection-error}

Our description of the scientific problem in Section \ref{subsec::introduction::problem-set-up} implicitly makes an important assumption:
\begin{assumption}
    \label{Assumption::A-3::prediction-not-imputation}
    The index of the entry to be predicted $(i_0,j_0)$ does not depend on $A$.
\end{assumption}

This issue concerned by Assumption \ref{Assumption::A-3::prediction-not-imputation} generally does not exist in the classical regression setting, where we always predict the \emph{next} response, indexed $n+1$.
However, in the matrix setting, it might be tempting to seek recovering \emph{missing} entries, a problem known as \emph{matrix imputation}, also referred to as \emph{matrix completion} \citep{candes2010matrix} or \emph{link prediction} \citep{zhao2017link}.
The key difference is that matrix imputation predicts an entry, not because it is the \emph{next} piece of data, but because it is \emph{missing}, following an unknown missing pattern.
To avoid potential confusions due to name conflicts, particularly with ``link prediction'', we stress that {\bf matrix prediction and matrix imputation are different problems}.

Readers may naturally wonder: is it possible to \emph{impute} missing entries distribution-free, only enforcing Assumptions \ref{Assumption::A-1::regularity-conditions} and \ref{Assumption::A-2::matrix-exchangeability}?
Unfortunately, the answer is ``No''.
The intuitive reason is that if $(i_0,j_0)$ may depend on $A$, then such dependence could render $(i_0,j_0)$ a \emph{random} index, and $A_{i_0,j_0}$ may no longer be row/column-exchangeable with other $A_{i,j}$'s.
It is crucial not to conflate this with Assumption \ref{Assumption::A-2::matrix-exchangeability}.
To clarify the concept, let us draw an analogy from the regression setting described in Section \ref{subsec::introduction::classical-conformal-prediction}, additionally assuming that ${\cal F}_{{\cal X}\times {\cal Y}}$ is smooth.
Instead of always missing the \emph{next} $Y$ (i.e. $Y_{n+1}$), we are now missing the \emph{largest} $Y$ (i.e., $Y_{i_0}$, where $i_0:=\arg\max_i Y_i$).
Then swapping $(X_{i_0}, Y_{i_0})$ and $(X_1,Y_1)$ in the ordered sequence $\big( (X_1,Y_1),\ldots,(X_{n+1},Y_{n+1}) \big)$ will alter the joint distribution, as the largest $Y$ will be invariantly $Y_1$ and may no longer appear at any index as before.
This situation is conceptually akin to \emph{Inspection Paradox} in stochastic processes.

Now, we formally show that our Assumption \ref{Assumption::A-3::prediction-not-imputation} cannot be waived.

\begin{theorem}[{\bf Impossibility of conformalized matrix imputation under arbitrary missingness}]
    \label{theorem::impossibility-theorem::no-nontrivial-conformal-imputation}
    Under Assumptions \ref{Assumption::A-1::regularity-conditions} and \ref{Assumption::A-2::matrix-exchangeability}, if the index of the entry to be predicted, i.e. $(i_0,j_0)$, may depend on the values of $A$'s entries (missing not at random), then the only conformal prediction method that guarantees $\pr(A_{i_0,j_0}\in {\cal I}_{i_0,j_0})\geq 1-\alpha$ for any distribution of $A$ and any missing mechanism $M$ is to always output the trivial prediction set $[-C_0,C_0]$.
\end{theorem}

\begin{remark}
    \label{remark::understandings::matrix-prediction-vs-matrix-completion}
    Theorem \ref{theorem::impossibility-theorem::no-nontrivial-conformal-imputation} remains true even when $(i_0,j_0)$ is the only missing entry.
    Hence, our paper exclusively focuses on \emph{matrix prediction} for a \emph{fixed} entry index $(i_0,j_0)=(n,n+1)$, perceived as the \emph{next data}.
    Aside from $(i_0,j_0)$, we allow arbitrary missing pattern elsewhere in the observed matrix.
    On the other hand, if we additionally assume \emph{missing completely at random (MCAR)} \citep{bland2015introduction}, the difference between matrix prediction and matrix imputation no longer exists. 
    In this case, the methods we develop in this paper can also serve as 
    valid conformalized matrix imputation methods.
\end{remark}

\begin{remark}
    \label{remark::global-assumptions}
    From this point onward, we will universally impose Assumptions \ref{Assumption::A-1::regularity-conditions}, \ref{Assumption::A-2::matrix-exchangeability}, and \ref{Assumption::A-3::prediction-not-imputation} in all theoretical results and will no longer repetitively cite them as prerequisites throughout.
\end{remark}

\subsection{Three types of coverage validity}
\label{subsec::understanding::three-types-validity}

In the classical regression setting, data are indexed by a single label $i$.
Here, in the matrix setting, each observation is indexed by both $i$ and $j$.
As a result, we now have \emph{three} different types of conformal prediction validity, instead of two types (\emph{marginal} and \emph{conditional}) found in the classical setting.
Using the definition of $\xi_i$'s from Section \ref{subsec::introduction::the-matrix-setting}, we can formally state the three validity types as follows.
\begin{enumerate}
    \item 
    {\bf Marginal validity:}
    $\pr(A_{n+1, n}\in {\cal I}_{n+1, n})\geq 1-\alpha$.
    
    \item 
    {\bf Row-conditional validity\footnote{Since column-conditional validity can be defined exactly similarly,
    we will simply use ``row-conditional'' to represent both concepts throughout this paper for succinct narration.}:}
    $\pr(A_{n+1, n}\in {\cal I}_{n+1, n}|\xi_{n+1})\geq 1-\alpha$.

    \item 
    {\bf Row-column-conditional validity:}
    $\pr(A_{n+1, n}\in {\cal I}_{n+1, n}|\xi_n, \xi_{n+1})\geq 1-\alpha$.
\end{enumerate}
Using a proof strikingly similar to that of Theorem \ref{theorem::impossibility-theorem::no-nontrivial-conformal-imputation}, one can show that nontrivial conformal prediction with row-column-conditional validity is impossible without further structural assumptions.
This aligns with similar findings in the regression setting \citep{lei2018distribution}.
Meanwhile, any row-conditionally valid method is also marginally valid (marginalize out $\xi_{n+1}$).
Therefore, in this paper, we concentrate on \emph{row-conditional validity}
and defer some discussion on marginal validity to Supplemental Material.

\section{Our method and theoretical justifications}
\label{section::our-method}

\subsection{General framework for full conformal matrix prediction with row-conditional validity}
\label{subsec::our-method::general-framework-full-conformal}

As previously discussed in Section \ref{subsec::introduction::classical-conformal-prediction}, developing a valid conformal prediction method involves designing an appropriate nonconformity score, denoted by $S_{n+1;j}(\cdot)$, that fulfills proper symmetry condition.
In the matrix setting, the novel data structure and potential presence of multiple missing entries demand an innovative definition of the symmetry requirement on the non-conformity score.
\begin{definition}[Row-conditional symmetry]
    \label{definition::row-conditional-symmetry-for-non-conformity-score}
    A non-conformity score $S_{n+1;j}(\cdot), j\in[1:n]$ is called ``row-conditionally symmetric'', if for any permutation: $\mathring \pi: [1:n]\leftrightarrow [1:n]$ and arbitrary input $\mathfrak{A}\in\mathbb{R}^{(n+1)\times (n+1)}$, it always satisfies
    $
        S_{n+1;j}({\cal A})
        =
        S_{n+1;\mathring \pi(j)}
        \big(
            {\cal A}_{\mathring \pi, \mathring \pi}
        \big).
    $
\end{definition}

\begin{lemma}
    \label{lemma::row-exchangeability-for-nonconformity-scores}
    If $S_{n+1;j}(\cdot)$ is row-conditionally symmetric\footnote{Also recall Remark \ref{remark::global-assumptions}.}, then it is row-conditionally exchangeable:
    \begin{align}
        \big(
            S_{n+1;1}(A), 
            \ldots, 
            S_{n+1;n}(A)
        \big)
        \big|\xi_{n+1}
        \stackrel{d}= &~
        \big(
            S_{n+1;\pi(1)}(A), 
            \ldots, 
            S_{n+1;\pi(n)}(A)
        \big)
        \big|\xi_{n+1},
    \end{align}
    for all $\pi: [1:n]\leftrightarrow[1:n]$.
\end{lemma}
Notice that in Lemma \ref{lemma::row-exchangeability-for-nonconformity-scores}, the nonconformity score uses the \emph{complete} data $A$ as the input.
This enables full conformal prediction designed based on Lemma \ref{lemma::row-exchangeability-for-nonconformity-scores} to handle \emph{arbitrary} missing pattern.

Now we are ready to describe our full conformal prediction framework.
Let the capital $Z\in \mathbb{R}^{(n+1)\times (n+1)}$ denote the matrix containing all guessed missing values, in which, only $\{Z_{i,j}: M_{i,j}=1\}$ matter, while all other elements in $Z$ can be set arbitrarily.
Set the shorthand 
\begin{align}
        \tilde A(Z) := A\circ (1-M) + Z\circ M,
        \quad 
        \textrm{and}
        \quad
        \tilde A(Z;z) := \tilde A(Z), 
        \textrm{ except } \big\{\tilde A(Z;z)\big\}_{n+1,n} = z,
        \label{def::shorthand::tilde-A}
\end{align}

where $(1-M)_{i,j} := 1-M_{i,j}$ and $(B_1\circ B_2)_{i,j} := (B_1)_{i,j} (B_2)_{i,j}$. 
Define
\begin{align}
    {\cal I}_{n+1;{\cal M}}
    :=
    \Big\{
        Z:
        S_{n+1;n}\big(\tilde A(Z)\big)
        \leq 
        {\cal Q}_{1-\alpha}
        \Big(
            n^{-1}
                \sum_{j=1}^n
                \delta_{S_{n+1;j}\big(\tilde A(Z)\big)}
        \Big)
    \Big\},
    \label{our-method::row-conditional::full-2}
\end{align}
where $\delta_{x_0}(u):=\mathbbm{1}_{[u\leq x_0]}$ denotes a point probability mass at $x_0$.
Then projecting ${\cal I}_{n+1;{\cal M}}$ onto $(n+1,n)$ yields our row-conditional full conformal prediction set ${\cal I}_{n+1;n}$, as follows:
\begin{align}
    {\cal I}_{n+1;n}
    :=&~
    \{
        z: \exists Z\in {\cal I}_{n+1;{\cal M}}, \textrm{ s.t. }Z_{n+1,n} = z
    \}.
    \label{our-method::row-conditional::full-3}
\end{align}
\begin{theorem}
    \label{theorem::conditional-validity::full-conformal::multiple-missing}
    We have 
    $
        \pr(A_{n+1,n}\in {\cal I}_{n+1;n}|\xi_{n+1}=u) \geq 1-\alpha
    $ 
    for any given $u\in[0,1]$.
    If, additionally, $(n+1,n)$ is the only missing entry and the distribution of $A$ is continuous, then we also have
    $
        \pr(A_{n+1,n}\in {\cal I}_{n+1;n}|\xi_{n+1}=u) \leq 1-\alpha + n^{-1}.
    $
\end{theorem}
The lower bound in Theorem \ref{theorem::conditional-validity::full-conformal::multiple-missing} always holds, but when there are multiple missing entries, the upper bound (``anti over-coverage'' property \citep{lei2018distribution}) might not hold.
This is because two disjoint multi-dimensional sets 
${\cal I}_{n+1;{\cal M}}^{(1)}$
and
${\cal I}_{n+1;{\cal M}}^{(2)}$
could potentially have overlapping projections onto the entry $(n+1,n)$.
In other words, with multiple missingness, ${\cal I}_{n+1;n}$ might be conservative.

\subsection{Fast algorithms for practitioners}
\label{subsec::our-method::practical-algorithms}

\subsubsection{Algorithm 1: Multiple initial guesses + SVD}

A simple design of nonconformity score is to use a matrix estimation technique, e.g. SVD, as a proxy:
\begin{align}
        S_{n+1;j}^{\textrm{SVD}}(\tilde A) 
        := 
        \big|
            \tilde A_{n+1;j}
            -
            \check A_{n+1;j}\big( \tilde A \big)
        \big|,
        \label{nonconformity-score::version-SVD}
\end{align}
Here, the matrix estimation $\check A(B)$ should comply with the symmetry requirement in Definition \ref{definition::row-conditional-symmetry-for-non-conformity-score}.
We use a modified USVT \citet{chatterjee2015matrix} as described in \citet{zhang2017estimating}:
$
    \check A(B)
    :=
    U_B S_B V_B^T,
$
where $B \approx U_B S_B V_B^T$ is a $n^{1/3}$-thresholded SVD of $B$.
The next question is how to efficiently implement \eqref{our-method::row-conditional::full-2} and \eqref{our-method::row-conditional::full-3}.
The effective dimensionality of ${\cal I}_{n+1;{\cal M}}$ is $m_0:=\sum_{1\leq j<i\leq n}M_{i,j}$.  
When $m_0=1$ (i.e., only $(n+1,n)$ is missing), we can implement \eqref{our-method::row-conditional::full-2} and \eqref{our-method::row-conditional::full-3} through a one-dimensional grid search for $z := (\textrm{guessed }A_{n+1,n})$.
When $m_0$ is moderately large, exact implementation of \eqref{our-method::row-conditional::full-2} and \eqref{our-method::row-conditional::full-3} becomes infeasible, but we can break the computation into two steps: 
(i) fill in missing entries with an initial guess $\hat Z$, yielding $\tilde A(\hat Z;z)$; 
(ii) perform a one-dimensional grid search for $z$, which produces a prediction set ${\cal I}_{n+1;n}(\hat Z)$.
The exact full conformal involves exhausting all possible $\hat Z$'s; here, we emulate it by taking a union of ${\cal I}_{n+1;n}(\hat Z)$ over many initial guess $\hat Z$'s as our final prediction ${\cal I}_{n+1;n}$.

\begin{remark}
    \label{remark::practical-full-conformal}
    Strategies for guessing $\hat Z$ include: 
    entry-wise i.i.d. sampling from the empirical distribution of observed $A$ entries; 
    filling all with $C_0$;
    all with $-C_0$; or a mixture of $\pm C_0$.
\end{remark}

\begin{algorithm}
    \caption{Approximate full conformal matrix prediction using SVD}\label{algorithm::full-conformal-SVD}
        \textbf{Input:} 
        $M\in\{0,1\}^{(n+1)\times (n+1)}$; 
        $\{A_{i,j}: (i,j):M_{i,j}=0\}$;
        $(i_0,j_0)=(n+1,n)$;
        $\alpha$; IterMax.
        \\
        \textbf{Output:}
        Row-conditional $1-\alpha$ conformal prediction interval ${\cal I}_{n+1;n}$.
    \begin{algorithmic}
    \For{$\ell$ in 1:IterMax}
        \State 1. 
        Draw initial guess $\hat Z^{(\ell)}$ following Remark \ref{remark::practical-full-conformal} and compute $\tilde A(\hat Z^{(\ell)};z)$ using \eqref{def::shorthand::tilde-A}.
        \State 2. Perform one-dimensional full conformal prediction on $z$, with \eqref{nonconformity-score::version-SVD}:
        \begin{align}
            {\cal I}^{(\ell)}_{n+1;n}
            :=
            \Big\{
                z:
                S_{n+1;n}^{\textrm{SVD}}\big(\tilde A(\hat Z;z)\big)
                \leq &~
                {\cal Q}_{1-\alpha}
                \Big(
                    n^{-1}
                        \sum_{j=1}^n
                        \delta_{S_{n+1;j}^{\textrm{SVD}}\big(\tilde A(\hat Z;z)\big)}
                \Big)
            \Big\}.
            \label{our-method::row-conditional::approximate-full-conformal-SVD}
        \end{align}
    \EndFor
    \State Return: 
    $
        {\cal I}_{n+1;n}
        :=
        \cup_{\ell=1}^{\textrm{IterMax}} 
        {\cal I}^{(\ell)}_{n+1;n}.
    $
    \end{algorithmic}
\end{algorithm}

\subsubsection{Algorithm 2: One initial guess + Algorithmic Stability + Neighborhood Smoothing}

Our second algorithm approaches the full conformal prediction by decoupling the dependency of most conformity scores from $Z$.
The overarching idea of tracking the change of conformity scores with the candidate prediction can be found in \citet{lei2014distribution, lei2018distribution}, but these methods are designed  specifically for their regression settings.
A recent pioneering work \citet{ndiaye2022stable} employs \emph{algorithmic stability} \citep{bousquet2002stability, kim2021black} for conformalized regression.
In this paper, we present two significant extensions of  Ndiaye's method:
first, we adapt it for the matrix setting; second, we show that it is not only an acceleration strategy as emphasized in \citet{ndiaye2022stable}, but also a powerful tool for handling missingness of arbitrary pattern.

We will present our method without recapitulating \citet{ndiaye2022stable} due to page limit.
Our perspective differs slightly from \citet{ndiaye2022stable} in that we focus on using the idea for handling multiple missingness, rather than completely avoiding full conformal computation.
To understand our method,
let us momentarily use God's eye view and access the non-observable \emph{complete} data $A$. 
Then we can use $A$ as an oracle guess for $Z$ in \eqref{our-method::row-conditional::full-2} and run a one-dimensional full conformal \eqref{our-method::row-conditional::full-3}.
The resulting prediction set ${\cal I}_{n+1;n}$ would have finite-sample coverage validity guarantee, since the problem reduces to the case of \emph{only one missing entry}.
However, in reality we can never access the complete $A$.  So, what shall we do?

The answer is surprisingly simple: just fill in \emph{any} guess $\hat Z$ for missing entries; then account for the incurred approximation error.
Notice that in the first step, we now only need \emph{one} initial guess.
For the crucial second step, we introduce the central concept: \emph{algorithmic stability} \citep{ndiaye2022stable}.
\begin{definition}[Algorithmic stability bound]
    \label{definition::algorighmic-stability-bound}
    For all $j\in [1:n]$, define
    \begin{align}
        \tau_{n+1;j}
        :=&~
        \sup_{z}\sup_{Z^{(1)}, Z^{(2)}}
        \big|
            S_{n+1;j}(\tilde A\big(Z^{(1)};z)\big)
            -
            S_{n+1;j}(\tilde A\big(Z^{(2)};z)\big)
        \big|.
    \end{align}
\end{definition}
\begin{remark}[Understanding Definition \ref{definition::algorighmic-stability-bound}]
    First, $\tau_{n+1;j}$ takes a supremum over $Z$, but over observed entries in $A$.
    Using algorithmic stability entails upper-bounding the variation of $S$. 
    Allowing $\tau$ to depend on $A$ provides tighter upper bound.
    Second, it is important to note that $\tau$ accounts for the worst case, relying solely on observable information.
    On one hand, this safeguards the method's coverage validity against any $M$ pattern, any $A$ distribution, and any relationship between $A$ and $M$.
    On the other hand, however, $A$ and $M$ can significantly impact the magnitudes of $\tau$'s and (through $\tau$'s) impact prediction accuracy.
    Third, unlike Definition 3.1 in \citet{ndiaye2022stable}, here we define $\tau$ \emph{without} using a fitted value $\hat A_{i,j}$ as a proxy.
    This brings technical convenience to the derivation of $\tau$'s.
\end{remark}

Now we formulate our method.
By definition,
$
    S_{n+1;n}\big(\tilde A(A;z)\big)
    \geq 
    S_{n+1;n}\big(\tilde A(\hat Z;z)\big) - \tau_{n+1;n}
$
and
$
    S_{n+1;j}\big(\tilde A(A;z)\big)
    \leq
    S_{n+1;j}\big(\tilde A(\hat Z;z)\big) + \tau_{n+1;j}.
$
Our prediction set is
\begin{align}
    {\cal I}^{(\rm{AS})}_{n+1;n}(\hat Z)
    :=
    \Big\{
        z: S_{n+1;n}\big(\tilde A(\hat Z;z)\big) - \tau_{n+1;n}
        \leq
        {\cal Q}_{1-\alpha}
        \Big(
            n^{-1}
                \sum_{j=1}^n
                \delta_{S_{n+1;j}\big(\tilde A(\hat Z;z)\big) + \tau_{n+1;j}}
        \Big)
    \Big\}.
    \label{our-method::row-conditional::algorithmic-stability}
\end{align}
\begin{theorem}
    \label{theorem::conditional-validity::algorithmic-stability}
    For any $\hat Z$, we have
    $
        \pr(A_{n+1,n}\in {\cal I}^{(\rm{AS})}_{n+1;n}(\hat Z)|\xi_{n+1}=u)
        \geq 
        1-\alpha,
    $
    for all $u\in[0,1]$.
\end{theorem}
To run our method in practice, it remains to specify $S$ and its $\tau$.
When choosing $S$, the ease of deriving a meaningful $\tau$ is an important consideration.
Therefore, rather than \eqref{nonconformity-score::version-SVD}, we propose an alternative choice for $S$.
This choice was inspired by the \emph{neighborhood smoothing} method \citep{zhang2017estimating}:
\begin{align}
    K_h(u) := \max(1-|u|/&h,0);\ \ 
    k_{h;j,j'}({\cal A})
    := 
    K_h\Big(
        \{n(n-2)\}^{-1}\sum_{\ell\neq i,i'} 
        \big|
            \langle 
                {\cal A}_{\cdot,j} 
                - 
                {\cal A}_{\cdot,j'}, 
                {\cal A}_{\cdot, \ell} 
            \rangle
        \big|
    \Big);
    \label{def::kernel-function}
    \\
    \textrm{and}\quad
    S_{n+1;j}\big( \tilde A(\hat Z) \big)
    := &~
    \sum_{j'\in [1:n]\backslash\{j\}}
    k_{h_j; j,j'}\big(\{\tilde A(\hat Z)\}_{1:n,1:n}\big) 
    \Big|
        \{\tilde A(\hat Z)\}_{n+1,j}
        -
        \{\tilde A(\hat Z)\}_{n+1,j'}
    \Big|,
    \label{our-method::row-conditional::full-1}
\end{align}
where ${\cal A}\in\mathbb{R}^{n\times n}$; $k_{h_j;j,j'}(\cdot)$ measures the structural dissimilarity between columns $j$ and $j'$; and $h_j$ is the kernel bandwidth, adaptively tuned for $j$, see \citet{zhang2017estimating, li2020network}.
In \eqref{def::kernel-function}, we define $k$ differently from \citet{zhang2017estimating} for analytical convenience.
With the choice of $S$ specified by \eqref{def::kernel-function} and \eqref{our-method::row-conditional::full-1}, we can derive $\tau$.

\begin{theorem}[Algorithmic stability bound for \eqref{def::kernel-function} and \eqref{our-method::row-conditional::full-1}]
    \label{theorem::stability-bound::row-conditional}
    Suppose $K_1(\cdot)$ is $L_0$-Lipschitz.  Set
    \begin{align}
        \tau_{n+1;j}
        := &~
        \min\big\{
            4L_0 C_0^3 h^{-1}\cdot 
            \big(
                m_j + 3\bar m
            \big),
            2C_0C_k
        \big\}
            +
            2C_KC_0\big\{
                (n-2)M_{n+1,j} + m_{n+1}
            \big\},
        \label{eqn::stability-bound::row-conditional}
    \end{align}
    where $m_i:=\sum_{j:j\neq i}M_{i,j}$ denotes the number of missing entries in row $i$,
    and $\bar m:=n^{-1}\sum_{j=1}^n m_j$,
    and $C_K:=\sup_u K_h(u), \forall h>0$.
    Then \eqref{eqn::stability-bound::row-conditional} would be a valid choice for $\tau_{n+1;j}$.
\end{theorem}

Now we are ready to present our second practical algorithm:

\begin{algorithm}[h!]
\caption{Accelerated full conformal matrix prediction using algorithmic stability}\label{algorithm::algorithmic-stability}
    \textbf{Input:} 
    $M$; 
    $\{A_{i,j}: M_{i,j}=0\}$;
    $(i_0,j_0)=(n+1,n)$;
    $\alpha$; $h$; $\hat Z$; $\{\tau_{n+1;j},j\in[1:n]\}$ by \eqref{eqn::stability-bound::row-conditional}.
    \\
    \textbf{Output:}
    Row-conditional $1-\alpha$ conformal prediction interval for $A_{n+1,n}$, denoted by ${\cal I}_{n+1;n}^{\textrm{(AS)}}(\hat Z)$.
\begin{algorithmic}
    \State Initialize $D = 0^{n\times n}$, then set 
    $
    D_{j,j'}
    \gets
    \big|
        \{\tilde A(\hat Z)\}_{n+1,j}
        -
        \{\tilde A(\hat Z)\}_{n+1,j'}
    \big|, \ 
    \forall j,j'\in [1:(n-1)]$;
    \For{each candidate $z$ in the search grid}
    \State Set 
    $
    D_{n,j'}
    =
    D_{j',n}
    \gets
    \big|
        z-
        \{\tilde A(\hat Z;z)\}_{n+1,j'}
    \big|, \ 
    \forall j'\in [1:n]$;
    \State Compute $S_{n+1;j}\big( \tilde A(\hat Z;z) \big)
    = 
     \sum_{j'\in [1:n]\backslash\{j\}}
    k_{h_j; j,j'}\big(\{\tilde A(\hat Z)\}_{1:n,1:n}\big)\cdot D_{j,j'}, \ 
    \forall j\in [1:n]$,
    \State Determine whether to include this $z$ in ${\cal I}_{n+1;n}^{\textrm{(AS)}}(\hat Z)$ according to \eqref{our-method::row-conditional::algorithmic-stability}.
    \EndFor
\end{algorithmic}
\end{algorithm}
In Algorithm \ref{algorithm::algorithmic-stability}, we used the fact that  $D_{[1:(n-1)], [1:(n-1)]}$ does not involve $z$ to speed up computation.

Theorem \ref{theorem::stability-bound::row-conditional} quantifies the impact of the proportion of missing entries on prediction accuracy via $\tau$.
We draw two observations.
First, missingness anywhere in the matrix, even outside row $n+1$, can affect $\tau$.
To see this, notice that \eqref{def::kernel-function} and \eqref{our-method::row-conditional::full-1} effectively construct artificial \emph{predictor-response} pairs, and any missing entry can influence either the predictor or response.
Then our second observation naturally follows: missingness in row $n+1$ has a more substantive impact on prediction accuracy, since this row contains the \emph{responses}.
This intuition is also quantitatively reflected in \eqref{eqn::stability-bound::row-conditional}, where $m_{n+1}$ appears in all $\tau_{n+1;j}$'s expressions.

If at least $\alpha$ proportion of entries are missing in row $n+1$, Algorithm \ref{algorithm::algorithmic-stability} will output the trivial prediction $[-C_0,C_0]$.
Is this our method's limitation or a fundamental limit?
We answer it up next.

\subsection{Fundamental limit}
\label{subsec::our-method::fundamental-limits}

\begin{theorem}[Fundamental Limit Theorem on proportion of missing entries]
    \label{theorem::impossibility-theorem}
    In either of the following scenarios:
    \begin{enumerate}[(a)]
        \item missing at least $\lceil \alpha \rceil$ proportion of entries in $A$, and we aim at marginal coverage validity,
        \item missing at least $\lceil \alpha \rceil$ proportion of entries in row $n+1$, and we aim at row-conditional coverage validity,
    \end{enumerate}
    the only conformal prediction method that ensures $1-\alpha$ coverage for all underlying distributions of $A$ and missing mechanisms $M$ is to invariantly output the trivial prediction set $[-C_0,C_0]$.
\end{theorem}

Theorem \ref{theorem::impossibility-theorem} suggests that achieving marginal validity requires a weaker assumption on missingness than row-conditional validity.
This is not surprising as the latter validity implies the former.
Due to page limit, we relegate a more detailed discussion about this to Supplemental Material.

\section{Simulations and data example}
\label{section::simulations-and-data}

{\bf Simulation set up.}
We consider three graphons (recall Section \ref{subsec::introduction::the-matrix-setting}) for data generation:
$
    f_1(u,v) := 5/2\cdot(u + v) - 0.75
$
(low-rank and smooth);
$
    f_2(u,v) := 5/2\cdot\cos[
        0.1/\{(u-1/2)^3+(v-1/2)^3+0.01\}]
    \cdot\text{max}(u,v)^{2/3}+2
$
(high-rank);
and
$
    f_3(u, v) := 5/3\cdot(u^2 + v^2)\cdot \cos \{(u^4 + v^4)^{-1}\}+0.75,
$ 
which emulates a non-smooth graphon, respectively.
We generate latent row/column positions by
$
    \xi_1,\ldots,\xi_n\stackrel{\rm i.i.d.}\sim$~Uniform$(0,1);
$
and we will manually vary $\xi_{n+1}\in\{0.1,0.2,\ldots,0.9\}$ as different $\xi_{n+1}$'s may pose varying difficulty levels.
For $f\in\{f_1,f_2,f_3\}$, the complete $A$ is given by
$
    A_{i,j}
    =
    f(\xi_i,\xi_j) + e_{i,j},
$
where $e_{i,j}=e_{j,i}\stackrel{\rm i.i.d.}\sim$~Uniform$(-0.1,0.1)$.
The goal is to predict $A_{n+1,n}$ with a $1-\alpha=90\%$ conditional (on $\xi_{n+1}$) confidence.
We compare our two algorithms with four benchmarks:
(i) {\tt missMDA} \citep{josse2016missmda},
(ii) {\tt softImpute} \citep{mazumder2010spectral},
(iii) {\tt mice} \citep{vanBuuren2011a},
and
(iv) {\tt PPCA} \citep{sportisse2020estimation}.
Among them, (i) and (ii) are matrix decomposition methods, where (ii) is a nuclear norm penalized variant of (i);
(iii) and (iv) are Bayesian methods: (iii) uses Gibbs sampling to perform multiple imputation, and (iv) leverages probabilistic PCA.
Method (iv) naturally outputs a CI, and methods (i)--(iii) provide or are compatible with multiple imputations, from which, we can produce CI's from the empirical distributions of their outputs.
We varied network size $n \in \{50, 100, 200, 400\}$ ($n=100, 400$ shown in Supplemental Material), but {\tt mice} and {\tt PPCA} would timeout ($\geq 12$ hours) for $n\geq 100$. 
We repeated the experiments 1000 times for our methods and {\tt softImpute}, and 300 times for the other benchmarks due to their speed constraints. 
We compare the performance of all methods in three aspects:
(i) \emph{empirical row-conditional coverage probability}, 
(ii) \emph{CI length}, and (iii) \emph{time cost}. 

\begin{figure}[h!]
    \centering
    \makebox[\textwidth][c]{
     \includegraphics[width=\textwidth]{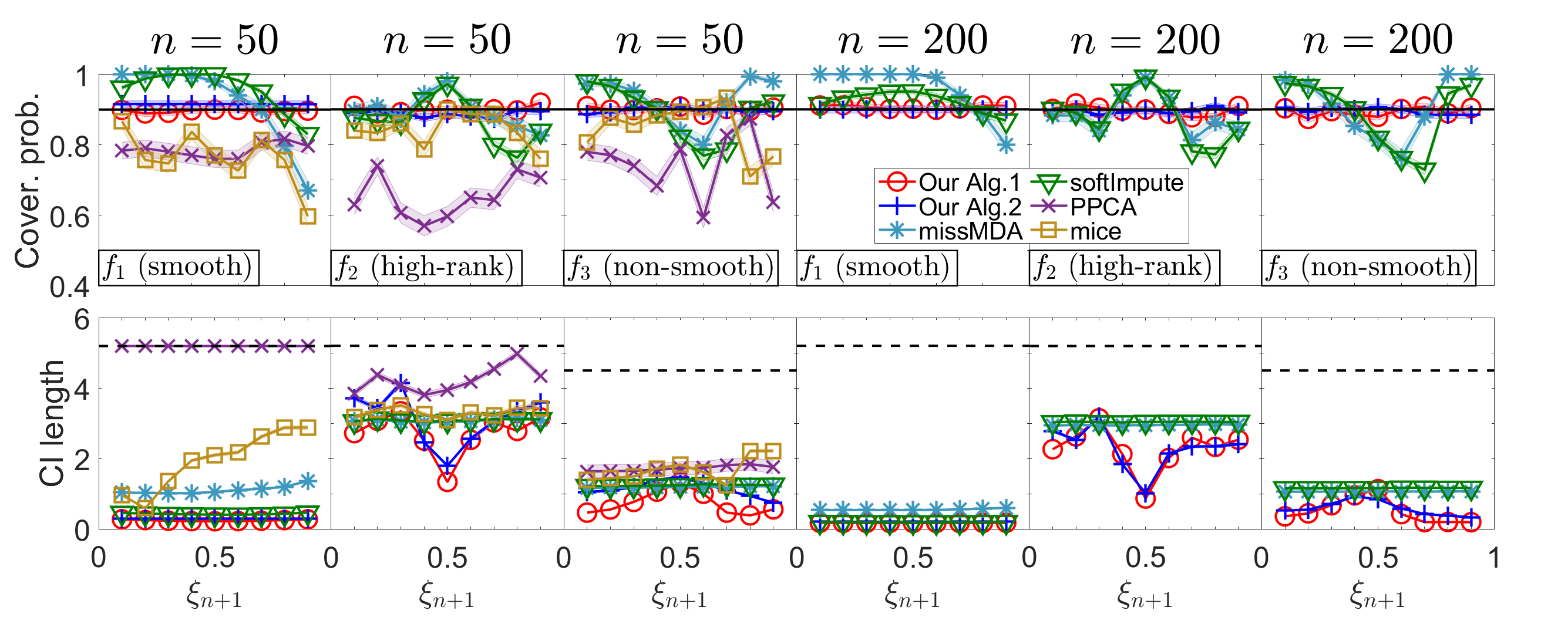}
    }    
    \caption{Performance comparison: missing $A_{n+1,n}$ only. 
    Columns indexed by data generation models.
    Black solid line (row 1)~$= 90\%$; 
    black dashed line (row 2)~$=2C_0$ (length of trivial CI). 
    }
    \label{figure::simulation1::missing-one-entry-performance}
\end{figure}

{\bf Simulation 1.}
This simulation focuses on the \emph{``distribution-free''} aspect of our method.
We set $M_{n+1,n}=M_{n,n+1}=1$ and $M_{i,j}=0$ elsewhere.
Figure \ref{figure::simulation1::missing-one-entry-performance} demonstrates that our algorithms are fast and maintain consistent row-conditional validity across all $\xi_{n+1}$'s under all graphons, while all benchmarks experienced varying levels of difficulty in maintaining coverage validity. 
Our methods also produce competitively short CI lengths. 
Also, our methods tend to produce longer CI's in settings where other methods struggled with coverage validity, and shorter CI's for easier cases (see, for example, column: $f_2$ \& $n=50$).
Since we did not introduce multiple missingness in this simulation, the observed advantage of our methods can be attributed to their \emph{conformal} nature.

\begin{figure}[ht]
    \centering
    \makebox[\textwidth][c]{
     \includegraphics[width=\textwidth]{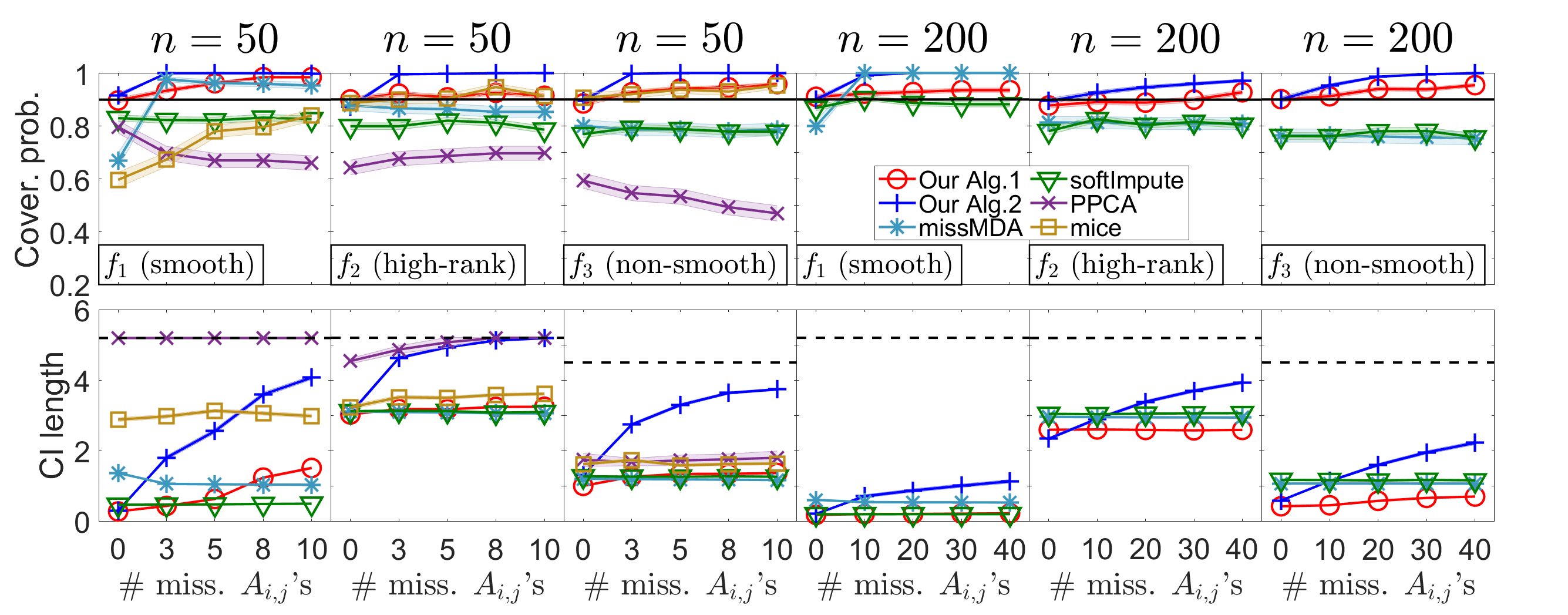}
    }
    
    \caption{Performance comparison: multiple-missingness. 
    Columns indexed by data generation models.
    Black solid line (row 1)~$= 90\%$; 
    black dashed line (row 2)~$=2C_0$ (length of trivial CI). 
    }
    \label{figure::simulation2::missing-multiple-entry-performance}
\end{figure}

\begin{figure}[ht]
    \centering
    \makebox[\textwidth][c]{
     \includegraphics[width=0.3\textwidth]{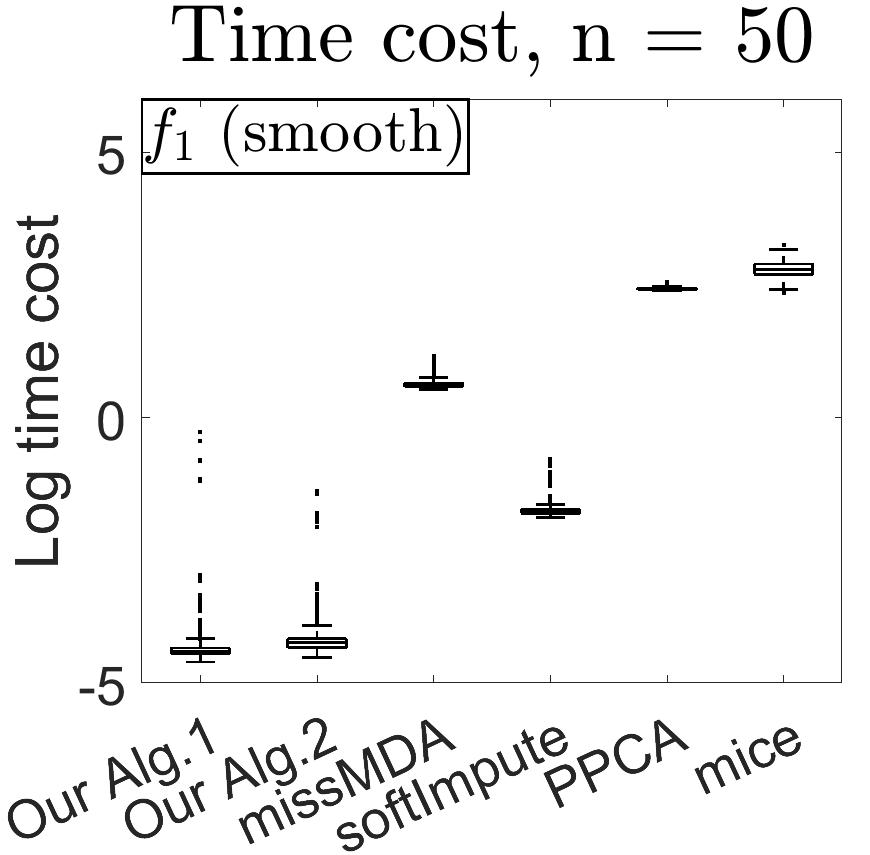}
    \includegraphics[width=0.3\textwidth]{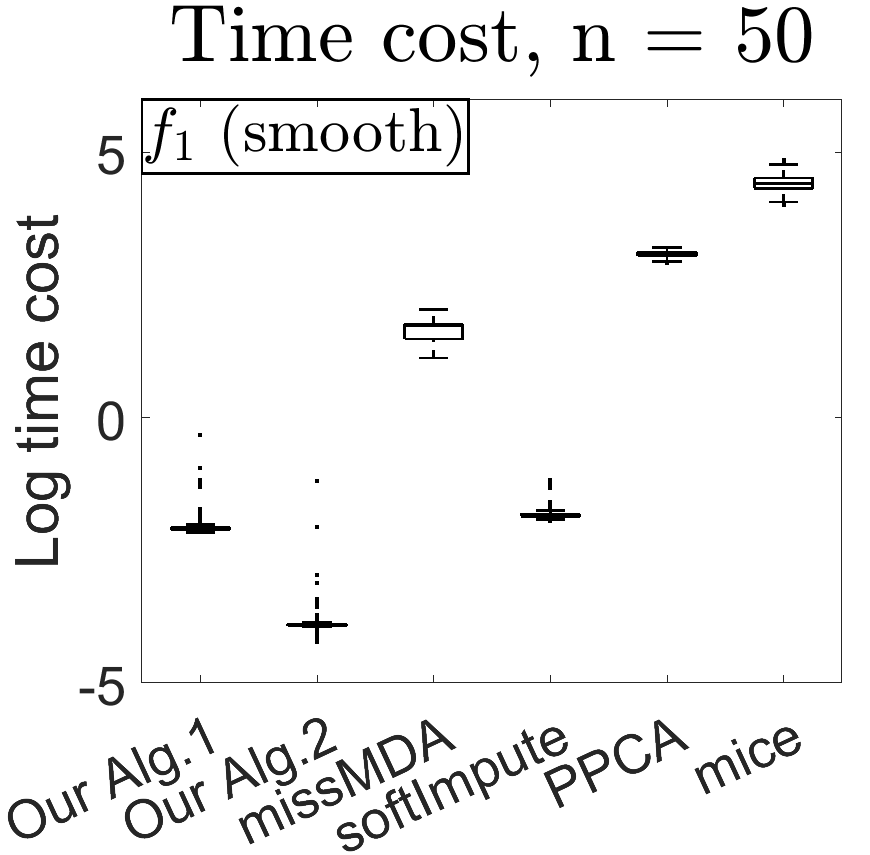}
    }
    
    \caption{Boxplots of computation time costs under graphon $f_1$ simulations. 
    Left: Simulation 1; 
    right: Simulation 2.
    Y-axis records \emph{log time cost}.
    Results for other graphons in Supplementary Material.
    }
    \label{figure::simulation2::missing-multiple-entry-performance}
\end{figure}

{\bf Simulation 2.}
This simulation examines the impact of \emph{multiple missingness}.
We largely inherited the data generation mechanism from Simulation 1 but fixed $\xi_{n+1} = 0.9, 0.7, 0.6$ for the three graphons, respectively, for clean visualization.
We designated a varying number of \emph{largest} entries in $A$ as missing -- this is \emph{missing not at random (MNAR)}.
Figure \ref{figure::simulation2::missing-multiple-entry-performance} presents the result.
Our methods respond to increasing levels of multiple missingness by widening prediction intervals; and once again, they show a clear advantage in maintaining coverage validity over all settings.
Comparatively, Algorithm \ref{algorithm::full-conformal-SVD} shows superior accuracy and speed;
Algorithm \ref{algorithm::algorithmic-stability}, although supported by rigorous theory, may sometimes appear conservative.

\begin{figure}
    \centering
     \includegraphics[width=0.8\textwidth]{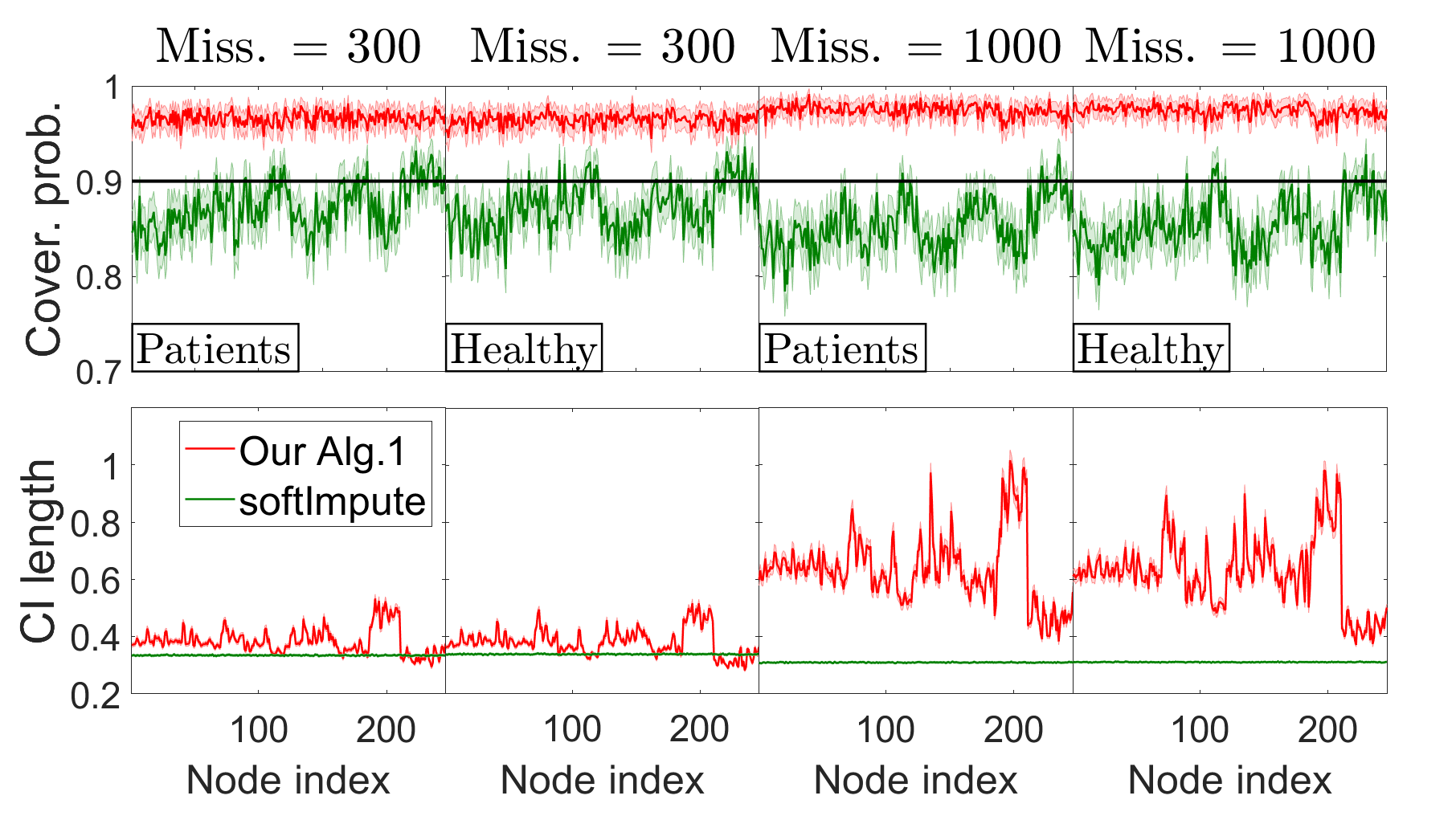}
   \caption{Performance comparison on schizophrenia data.  Algorithm \ref{algorithm::algorithmic-stability} produced trivial prediction.
   Benchmarks other than {\tt softImpute} timed out (timeout~$= 12$~hours).
    }
    \label{figure::data-example-missing}
\end{figure}

{\bf Data example.}
The data set \citet{adhikari2019functional} contains functional brain connectivity networks on 246 shared nodes among 103 schizophrenia patients and 124 healthy controls. 
Each entry represents a transformed correlation between the blood oxygen levels of two terminal nodes.
We assume that the networks in the two groups follow their respective population distributions. 
To assess row-conditional validity, such as for row 1 in the first patient's network, we randomly selected an off-diagonal entry as the prediction target and apply all methods. 
We replicated this process 8 times for all individuals, resulting in a total of $103\times 8=824$ binary outcomes (``cover'' or ``no-cover'') for each method and each row in the patient group; we computed similarly for the healthy group. 
In each replication, we randomly selected the largest $m_0\in\{300, 1000\}$ entries in each matrix to be missing.
We mainly compared our Algorithm \ref{algorithm::full-conformal-SVD} to {\tt softImpute}, as  {\tt missMDA}, {\tt mice} and {\tt PPCA} all timed out; and Algorithm \ref{algorithm::algorithmic-stability} outputted $[-C_0,C_0]$.
Figure \ref{figure::data-example-missing} shows the result.
In comparison to benchmarks, our Algorithm \ref{algorithm::full-conformal-SVD} demonstrates clear advantages in both coverage validity (conditional on any row) and computation speed (see Table \ref{table::data-example::time-comparison}).
Noticeably, our Algorithm \ref{algorithm::full-conformal-SVD} successfully safeguarded row-conditional validity against an adversarial missing pattern designed to introduce systematic biases. 
We can clearly observe increased CI lengths in our method in rows where the coverage validity of {\tt softImpute} dropped most significantly, which also suggests that prediction problem is more challenging for those rows.

\begin{table}[ht]
    \centering
    \caption{Computation time comparison (in seconds) in the data example, $m_0 = 300$. Mean(std. dev.)}
    \label{table::data-example::time-comparison}
    \begin{tabular}{|c|c|c|c|c|}
        \hline 
           &\multicolumn{2}{c|}{Patients} & \multicolumn{2}{c|}{Healthy controls} \\
            \hline
        Method & Our Algorithm \ref{algorithm::full-conformal-SVD} & {\tt softImpute} & Our Algorithm \ref{algorithm::full-conformal-SVD} & {\tt softImpute}  \\\hline
        Time cost & 0.212(0.018) & 1.349(0.144) & 0.208(0.012) & 1.346(0.125)  \\\hline
    \end{tabular}

\end{table}

\section{Discussion}
\label{section::discussion}

We briefly compare the two algorithms proposed in this paper: Algorithm \ref{algorithm::full-conformal-SVD} is conceptually simple and decently fast, but it requires multiple initialization and lacks theoretical guarantee as an approximation method;
Algorithm \ref{algorithm::algorithmic-stability} is fast as it only requires one initialization and enjoys provable finite-sample coverage validity, but it may tend to produce conservative prediction intervals.
Addressing these limitations would be an interesting future work -- specifically, pursuing a better understandings of the finite-sample theoretical properties of Algorithm \ref{algorithm::full-conformal-SVD}; and exploring improved algorithmic stability bounds for Algorithm \ref{algorithm::algorithmic-stability} to reduce its conservativeness.

As aforementioned, Assumption \ref{Assumption::A-1::regularity-conditions} is largely non-essential, and its relaxations could lead to various similar variants of the methods presented in this paper.
First, the universal boundedness assumption does not limit generality, as we can always consider $\varsigma(A) := \{\varsigma(A_{i,j}), 1\leq \{i,j\}\leq n\}$, where $\varsigma: \mathbb{R}\leftrightarrow (-C_0, C_0)$ is a monotone and bijective function \citep{rizzo2016energy}.
Then the link prediction output for the missing entries in $\varsigma(A)$ can be easily mapped back using $\varsigma^{-1}$ -- this produces predictions for the missing entries in $A$, in their original scales while preserving coverage guarantees.
Second, our method does not rely on the assumption that $A$ is symmetric
-- it can be readily adapted for asymmetric $A$'s, such as directed networks or bipartite graphs.
The main additional work is to differentiate between row- and column-conditional validity; 
and in the event of the latter request, apply our method onto $A^T$ instead of $A$.
A new fundamental limit result, as a slight variation of Theorem \ref{theorem::impossibility-theorem}, will govern the conformal prediction procedure.




\bibliographystyle{apalike}
\bibliography{all-ref}

\end{document}